\documentclass[upint,subscriptcorrection,varvw,hyphenate,balance,greek,russian,vietnamese,german]{asmeconf} 

\usepackage{makecell}
\usepackage{tablefootnote}
\usepackage{pifont}
\usepackage{float}

\hypersetup{%
	pdfauthor={Michael Holm},									  
	pdftitle={Lowering the Barrier to AI-Driven Inspection: A No-Code Workflow for Automated Structural Defect Detection},                  
	pdfkeywords={ASME conference paper, computer vision, defect detection, structural health, open source},
	pdfsubject = {Computer vision tool for structural health monitoring},			  
}

\allowdisplaybreaks 

\begin{document}

\setlength{\headwidth}{\textwidth}


\ConfName{Proceedings of the ASME 2026 Premier Conference on\\
Smart Materials, Adaptive Structures and Intelligent Systems}
\ConfAcronym{SMASIS2026}
\ConfDate{September 13--16, 2026} 
\ConfCity{Albuquerque, NM}
\PaperNo{SMASIS2026-190654}

%

\title{Lowering the Barrier to AI-Driven Inspection: A No-Code Workflow for Automated Structural Defect Detection} 
 
%
%
%

\SetAuthors{%
    Michael Holm\affil{1}\CorrespondingAuthor{holm3@purdue.edu}, 
	Tanner McElroy\affil{1}, 
	Xinghang Zhang\affil{1}, 
	Guang Lin\affil{1},  
	}

\SetAffiliation{1}{Purdue University, West Lafayette, IN}


\maketitle




\begin{abstract}
Structural health monitoring (SHM) is essential in modern engineering, providing data for condition-based maintenance, lifecycle assessment, and predictive decision-making. Traditionally, SHM relied on visual inspection to detect defects such as cracks and deformations. Early computer vision (CV) methods, including thresholding, edge detection, and handcrafted features, aimed to automate this process but were highly sensitive to noise, imaging variations, and multiscale defects, limiting their reliability.

Recent advances in machine learning, particularly convolutional neural networks (CNNs) and You Only Look Once (YOLO), have improved defect detection accuracy and enabled real-time analysis. However, adoption in SHM remains limited due to technical barriers such as data labeling, model training, and deployment, which typically require programming expertise.

To address this gap, we introduce YOLOEZ, an open-source, GUI-based tool for end-to-end YOLO model application. YOLOEZ integrates data labeling, training, and inference into a single interface, enabling high-performance model development without code while supporting reproducible workflows.

Evaluation against existing software and classical image processing demonstrates that YOLOEZ not only outperforms traditional methods across most detection metrics, but also lowers adoption barriers present in other modern CV tools. By combining accuracy with accessibility, YOLOEZ facilitates wider use of AI-driven monitoring for predictive maintenance, digital twins, and intelligent structural systems.
\end{abstract}

\begin{center}
\fbox{\parbox{0.92\columnwidth}{\footnotesize
This is the authors' version of a paper accepted for publication in the
\emph{Proceedings of the ASME 2026 Conference on Smart Materials, Adaptive
Structures and Intelligent Systems (SMASIS2026)}, paper SMASIS2026-190654,
by Michael Holm, Tanner McElroy, Xinghang Zhang and Guang Lin.
ASME is the original publisher. Copyright \copyright{} 2026 by ASME.
Reproduced and distributed by the authors for non-commercial purposes under
the rights retained in the ASME Copyright Agreement.}}
\end{center}



\section{Introduction}
Structural health monitoring (SHM) plays a central role in modern engineering systems by enabling condition-based maintenance, lifecycle assessment, and predictive decision-making \cite{ZHANG2025116575}. As materials and structures become increasingly complex and performance-driven, there is a growing need for reliable, scalable methods to assess structural integrity over time \cite{LAMBINET2022110675}. Traditionally, SHM has relied heavily on visual inspection, where experts evaluate features such as crack density, crack propagation, and surface deformation \cite{https://doi.org/10.1111/str.12495}. While effective, these approaches are inherently time-consuming, subjective, and difficult to scale, particularly for high-resolution or large-volume datasets \cite{INSAIGLESIAS2021103755}.

To address these limitations, computer vision (CV) techniques have been introduced to automate the analysis of structural imagery \cite{doi:10.1177/1475921720935585}. Early methods primarily relied on classical image processing techniques, including thresholding, edge detection, and handcrafted feature extraction \cite{TSAI2019209}. Although computationally efficient and relatively simple to implement, these approaches are often sensitive to noise, imaging conditions, and the complex, multiscale characteristics of real-world defects. As a result, their reliability in practical SHM applications remains limited \cite{Tabernik2019SegmentationbasedDA}.

Recent advances in machine learning, particularly convolutional neural networks (CNNs) \cite{article}, have significantly enhanced the performance of defect detection and localization in structural imagery \cite{10589380}. Architectures such as You Only Look Once (YOLO) \cite{inproceedings} enable accurate, real-time detection, making them especially well-suited for automated inspection tasks. To facilitate the application of computer vision to structural health monitoring (SHM), a growing body of research has focused on developing effective methodologies and implementation strategies \cite{PENG2024118809, doi:10.1177/1475921720935585, JAYARAM2023}. This line of work highlights the technical challenges inherent in deploying such systems, including data labeling, model configuration, and inference pipeline development, tasks that often require specialized programming and machine learning expertise.

This gap between technological capability and practical accessibility highlights the need for tools that enable domain experts to leverage machine learning without requiring extensive technical backgrounds. In this work, we introduce YOLOEZ, an open-source, graphical user interface (GUI)-based software tool designed to streamline the end-to-end application of YOLO-based models for structural defect detection. YOLOEZ integrates data labeling, model training, and inference into a unified and intuitive workflow, eliminating the need for code-based interaction and lowering the barrier to entry for materials and structural engineers, and is designed with usability as a core value. 

In addition to improving accessibility, YOLOEZ is designed with a strong emphasis on reproducibility and sustainable research practices. By providing a consistent framework for dataset generation, model development, and inference, the platform facilitates transparent and repeatable experimentation. This is particularly important in SHM, where variability in data and methodology can hinder the comparability of results across studies. YOLOEZ is released as an open-source project, with full documentation and source code available online; a dedicated software paper describing the implementation is currently under preparation for submission to the Journal of Open Source Software (JOSS).

\section{Background}

The application of computer vision to structural health monitoring (SHM) has evolved significantly over the past two decades, driven by advances in both imaging technology and machine learning \cite{PARK2023}. Early efforts to automate defect detection relied heavily on classical image processing techniques, including thresholding, edge detection, region growing, and morphological operations \cite{6263264}. These approaches are attractive due to their low computational cost, interpretability, and ease of implementation. In controlled environments, such methods can produce reliable results, particularly when defects exhibit strong contrast relative to the background \cite{ALAKNANDA200629}.

However, real-world SHM scenarios rarely conform to such ideal conditions \cite{doi:10.1177/1475921718757405}. Structural defects such as cracks, corrosion, and delamination often exhibit complex geometries, varying scales, and low contrast against heterogeneous backgrounds. Imaging conditions further complicate the problem, as variations in lighting, surface texture, noise, and occlusion can significantly degrade performance. As a result, classical methods tend to require extensive manual tuning and often fail when used on images with irregular backgrounds \cite{Jahanshahi01122009}. These limitations have motivated the transition toward data-driven approaches that can learn robust feature representations directly from labeled data \cite{Alamuru_2024}.

The emergence of deep learning, and in particular CNNs \cite{article}, has transformed the landscape of computer vision-based defect detection. CNNs are capable of learning hierarchical feature representations that capture both low-level texture and high-level semantic information, enabling them to outperform traditional methods in complex visual tasks. Within this domain, object detection architectures have proven especially valuable for SHM applications, as they enable both localization and classification of defects within an image \cite{CHA2024105328}.

Among these architectures, the YOLO \cite{inproceedings} family of models has gained widespread adoption due to its balance of accuracy and computational efficiency. Unlike two-stage detectors, which first generate region proposals and then classify them, YOLO operates as a single-stage detector that predicts bounding boxes and class probabilities directly from the input image in a single forward pass. This design enables real-time inference while maintaining competitive detection performance, making YOLO particularly well-suited for large-scale inspection tasks, video-based monitoring, and edge deployment scenarios \cite{10.1007/978-3-031-39603-8_15}.

Despite these advantages, the practical application of YOLO models in SHM is not straightforward. The end-to-end workflow for developing a functional model involves several interdependent stages: data collection, annotation, dataset formatting, model configuration, training, validation, and deployment. Each stage introduces its own technical challenges \cite{KHAN2025}. For example, data annotation requires careful labeling of defects, often with domain-specific ambiguity; training requires selecting appropriate hyperparameters and managing computational resources; and deployment requires constructing inference pipelines that integrate preprocessing, model execution, and postprocessing. These steps are typically implemented through code-based workflows, requiring familiarity with programming languages such as Python, machine learning frameworks, and command-line tools.

This complexity has created a gap between the capabilities of modern computer vision methods and their adoption in practice, particularly among domain experts in civil, mechanical, and materials engineering who may not have formal training in machine learning \cite{BAPPY2024}. As a result, there has been increasing interest in tools that aim to simplify and standardize the development pipeline for YOLO-based models.

One of the most influential tools in this space is Ultralytics \cite{jocher2023ultralytics}, which provides a widely used implementation of YOLO models along with a suite of utilities for training, validation, and inference. The Ultralytics framework has become a de facto standard for many practitioners due to its performance, flexibility, and active development \cite{10.1007/978-3-031-70285-3_48}. It offers pre-trained weights, configurable architectures, and support for custom datasets, allowing users to rapidly prototype and deploy models. Additionally, its integration with modern deep learning libraries enables efficient use of hardware accelerators such as GPUs.

However, despite these strengths, Ultralytics remains fundamentally oriented toward users with programming experience. Interaction with the framework typically occurs through Python scripts or command-line interfaces, requiring users to understand concepts such as dataset configuration files, training arguments, and model checkpoints. While documentation and community support have improved accessibility, the workflow still assumes a level of technical proficiency that may not be present among all SHM practitioners. Furthermore, tasks such as data annotation and dataset management are not natively integrated into the framework, requiring users to rely on external tools.

Complementing this ecosystem, Roboflow has emerged as a popular platform for dataset annotation, preprocessing, and management \cite{dwyer2026roboflow}. Roboflow provides a web-based graphical interface that allows users to label images, apply augmentations, and organize datasets with minimal technical overhead. Its emphasis on usability and collaboration has made it particularly attractive for teams working on computer vision projects. The platform also supports exporting datasets in formats compatible with YOLO and other frameworks, facilitating integration with downstream training pipelines.

While Roboflow significantly lowers the barrier to entry for dataset preparation, it does not provide a complete solution for model development. Training and inference are typically performed using external frameworks such as Ultralytics, requiring users to transition between platforms and manage intermediate artifacts. This separation introduces additional complexity, as users must ensure compatibility between dataset formats, model configurations, and evaluation procedures. Moreover, reliance on cloud-based services like Roboflow can raise concerns related to data privacy, cost, and long-term reproducibility, particularly in research and industrial contexts where data sensitivity is a consideration.

Dedicated annotation platforms such as CVAT \cite{cvat}, Label Studio \cite{labelstudio}, and VGG Image Annotator \cite{via} have also gained traction in the research community as open-source solutions for image labeling. These tools provide mature, feature-rich annotation interfaces that support a variety of label formats including bounding boxes, polygons, and segmentation masks. However, they are annotation-only platforms, choosing to leave model training, validation, and inference to other tools, reintroducing the workflow fragmentation described above. This separation requires users to manage dataset export formats, ensure compatibility with downstream training pipelines, and assemble their own evaluation procedures, tasks that present significant barriers for domain experts without programming experience.

The coexistence of these tools reflects a broader trend toward modular, loosely coupled
workflows that place a burden on users to integrate and manage multiple components.
In practice, this fragmentation can hinder reproducibility, as subtle differences in
preprocessing, training configuration, or evaluation metrics may lead to inconsistent
results, a concern especially pronounced in SHM, where datasets are often
heterogeneous and engineering decisions depend on the reliability and interpretability
of model outputs \cite{doi:10.1177/1475921720972416}. Consequently, there is a
growing need for integrated tools that unify the entire pipeline, from data labeling
to model deployment, within a single, coherent framework.

In this context, graphical user interface (GUI)-based systems offer a promising direction for improving accessibility and usability. By abstracting away low-level implementation details, such systems can enable users to focus on domain-specific tasks while still leveraging the power of advanced machine learning models. At the same time, a well-designed GUI can enforce standardized workflows, improving reproducibility and reducing the potential for error. However, existing solutions have yet to fully realize this vision, particularly in the context of YOLO-based object detection for SHM. Addressing this need, we developed YOLOEZ, a fully integrated, open-source, GUI-based environment for the end-to-end application of YOLO models in structural health monitoring.

\section{Methodology}

\begin{figure}[h]
\centering
\includegraphics[width=0.45\textwidth]{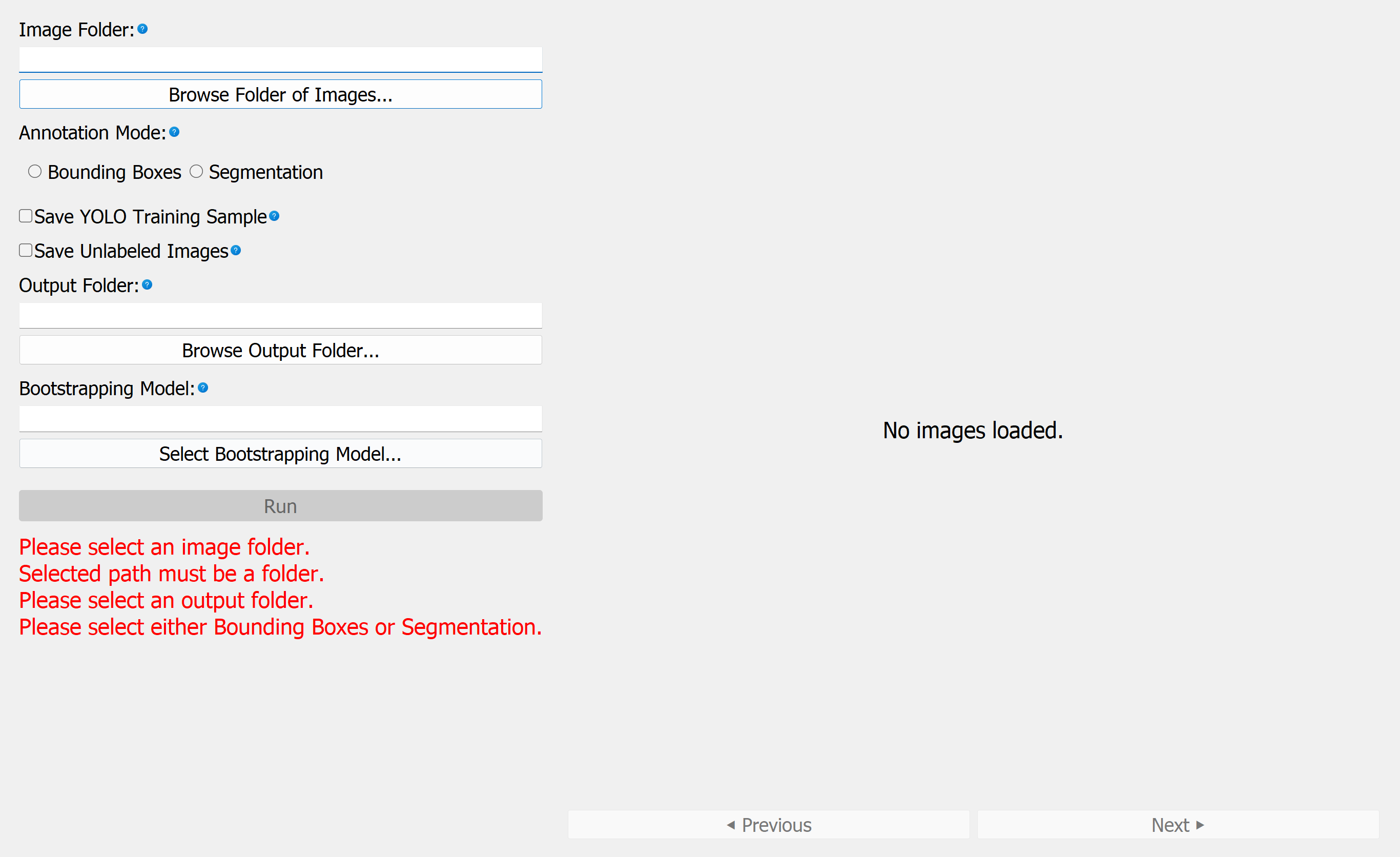}
\caption{The YOLOEZ graphical user interface, illustrating the tool's code-free interaction model. The interface consolidates data labeling, model training configuration, and inference into a single, integrated application, eliminating the need for command-line interaction or programming knowledge.}
\label{fig:yolo_gui}
\end{figure}

\begin{figure}[h]
\centering
\includegraphics[width=0.45\textwidth]{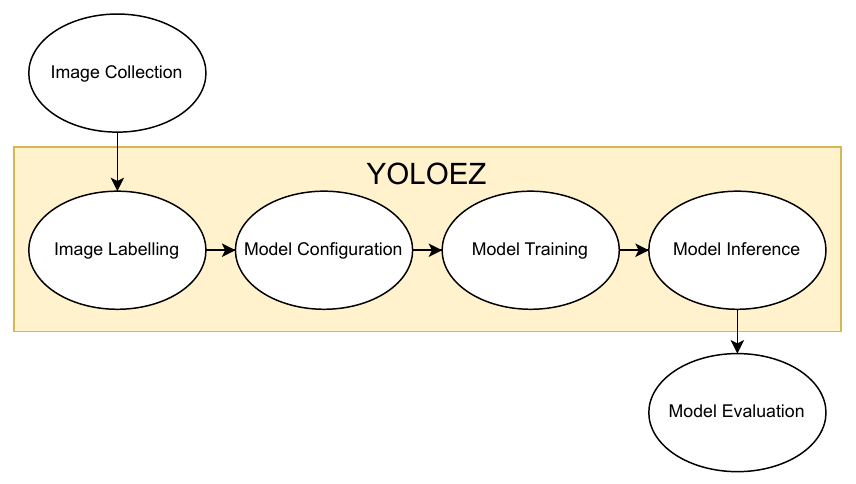}
\caption{End-to-end YOLOEZ workflow for YOLO-based defect detection. Starting from raw image data, the pipeline proceeds through image labeling, model configuration, model training, and inference, all managed within a unified graphical environment without requiring user-written code.}
\label{fig:yolo_workflow}
\end{figure}

To evaluate the effectiveness of an all-in-one, GUI-based tool for training and deploying YOLO models, YOLOEZ was developed in Python as a fully integrated platform for SHM applications. YOLOEZ consolidates data labeling, model training, and inference within a single graphical interface, eliminating the need for code-based interaction and reducing technical barriers for non-expert users. Shown in Fig.~\ref{fig:yolo_gui} is an example of one of YOLOEZ's GUI windows, and Fig.~\ref{fig:yolo_workflow} illustrates the intended end-to-end workflow, from image labeling to model inference.

This study evaluates YOLOEZ through both qualitative and quantitative analyses. The qualitative analysis examines workflow efficiency, usability, and accessibility compared to other computer vision frameworks. The quantitative analysis focuses on technical performance, specifically the detection and localization of micro-scale cracks in scanning electron microscope (SEM) images of 3D-printed tungsten. Together, these evaluations provide a comprehensive assessment of YOLOEZ, capturing both its effectiveness in defect detection and its ability to streamline the end-to-end SHM workflow.

\subsection{Qualitative Comparison of Software Tools}
To assess workflow efficiency and usability, YOLOEZ was compared to other prevalent computer vision frameworks, including Ultralytics, PyTorch-based YOLO implementations \cite{paszke2019pytorchimperativestylehighperformance}, and Roboflow. The comparison considered key workflow characteristics, including, but not limited to, the availability of a GUI, support for local data labeling, options for local and cloud-based training, inference capabilities, and overall workflow integration. This evaluation was based on these various tools' available documentation.

\subsection{Quantitative Evaluation: YOLOEZ versus Morphological Techniques}
\begin{figure}[h]
\centering
\includegraphics[width=0.45\textwidth]{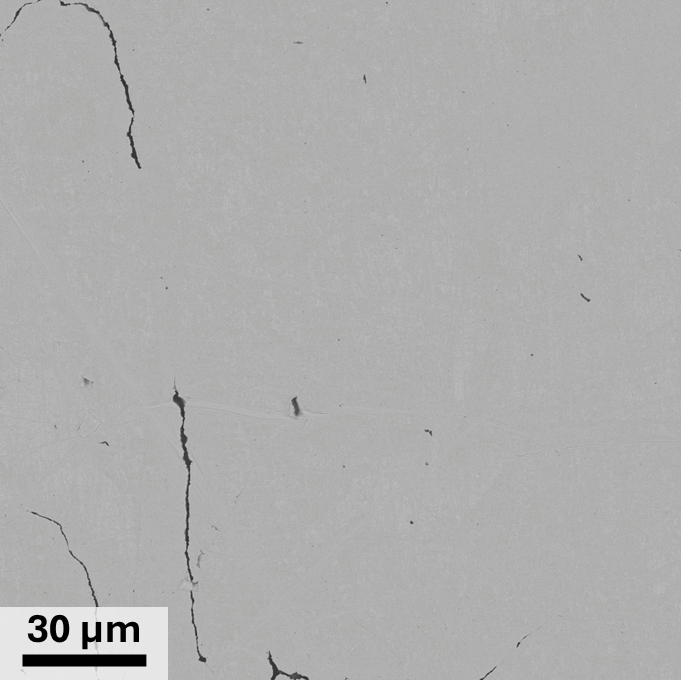}
\caption{Example scanning electron microscope image of additively manufactured tungsten showing micro-scale crack features. The image exhibits significant variability in crack morphology typical of challenging real-world SHM inspection scenarios.}
\label{fig:sem_image}
\end{figure}

To assess the technical performance of YOLOEZ, the tool was used to
fine-tune a pre-trained YOLO segmentation model, specifically a medium-sized YOLO11 segmentation model trained until convergence, on SEM images of
additively manufactured tungsten components containing micro-scale cracks
representative of real-world SHM defects; an example image is shown in
Fig.~\ref{fig:sem_image}. 

\begin{figure}[h]
\centering
\includegraphics[width=0.35\textwidth]{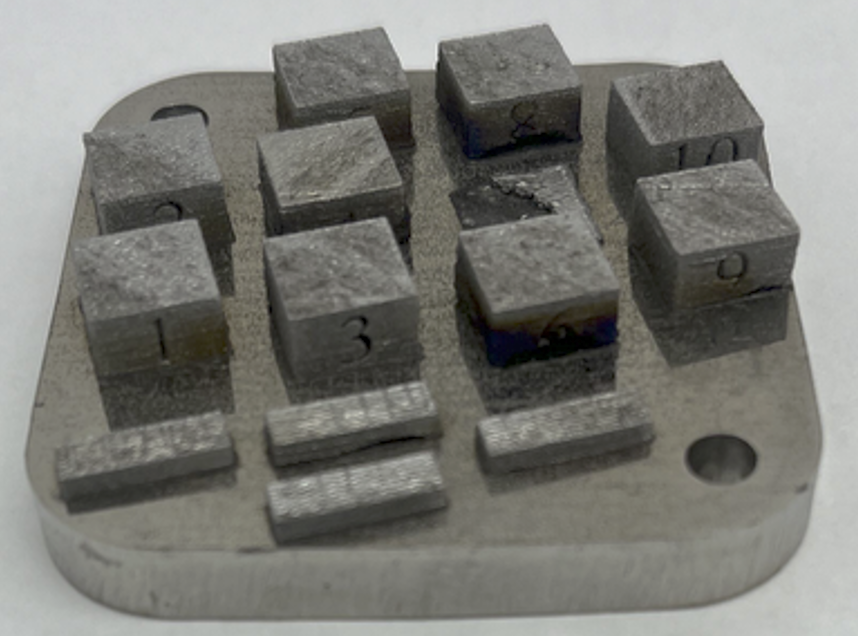}
\caption{Example additively manufactured tungsten components produced by laser powder bed fusion, showing the samples prior to sectioning and SEM imaging.}
\label{fig:component_raw}
\end{figure}

The test specimens were blocks of dimensions 8\,mm $\times$ 8\,mm $\times$ 5\,mm of pure tungsten produced via laser powder bed fusion (LPBF). During the LPBF process, tungsten powder is selectively melted by a high-power laser, layer by layer, to build the final part. Micro-scale cracks frequently develop in tungsten components due to the rapid heating and cooling cycles inherent in LPBF, which create high stresses and lead to crack nucleation at oxygen-embrittled grain boundaries, making them ideal candidates for evaluating defect detection algorithms. The components were prepared for imaging by cross-sectioning and polishing the surfaces to remove surface irregularities. Scanning electron microscopy was then used to capture high-resolution images of the polished surfaces, revealing the crack microstructure. Figure~\ref{fig:component_raw} shows an example of the as-built tungsten components before sectioning.

These images present several characteristics typical of challenging real-world inspection scenarios: heterogeneous surface texture that produces false contrast gradients, micro-scale crack widths that approach the resolution limit of the imaging system, and significant variability in crack morphology and density across samples. 

Images were partitioned into training,
validation, and test subsets of 20, 5, and 15 images, respectively,
reflecting the small labeled datasets typical of SHM research. To expand
the effective training and validation set, each of the 25 images was augmented
four times using YOLOEZ, yielding 100 additional synthetic samples. Each
augmented sample was generated by randomly applying a combination of
horizontal flipping, affine transformations with rotations of a magnitude of up to $30^{\circ}$ and scale factors between $0.8$ and $1.2$, color jitter,
Gaussian blur, and additive Gaussian noise. Ten of the 15 test images
were drawn from a separate 3D-printed tungsten sample to assess
out-of-distribution generalization. The inclusion of these images from a second, independently produced tungsten sample further reflects the specimen-to-specimen variability encountered in practical SHM deployments. Ground-truth crack annotations were
produced manually and served as the reference for all evaluations. To
account for the stochastic nature of model training, ten independent
models were trained and evaluated under this procedure, from which 95\%
confidence intervals on all performance metrics were derived. 

The in-distribution images used for this quantitative analysis will be publicly released on the Purdue University Research Repository alongside the in-progress paper titled `Crack mitigation in pure tungsten produced by laser powder bed fusion'. The out-of-distribution images will similarly be made available upon publication of a companion work currently in preparation, which focuses on studying mechanisms behind the evolution of micro-cracks in tungsten.

As a performance baseline, YOLOEZ was compared against a classical 
morphological image processing pipeline. To ensure a fair and reproducible comparison, the morphological baseline 
pipeline was implemented as a representative approach a domain expert 
without machine learning expertise might adopt. The pipeline consists of 
the following steps:

\begin{enumerate}
    \item \textbf{Preprocessing}: Convert the image to grayscale, normalize 
    intensity values to the range $[0, 255]$, and apply Gaussian blur with a 
    kernel size of $5 \times 5$ to reduce noise while preserving crack edges.
    
    \item \textbf{Thresholding}: Apply Otsu's automatic thresholding method 
    \cite{4310076}, which selects a threshold value that minimizes 
    the variance of the two resulting classes. The computed threshold is then 
    scaled by a factor of $1.35$ to create a slightly stricter threshold cutoff for crack detection. The result is inverted 
    to produce a binary mask where cracks appear as white pixels.
    
    \item \textbf{Morphological filtering}: Apply 1 iteration of morphological opening 
    followed by 12 iterations of closing using a $3 \times 3$ 
    circular structuring element. Opening removes small noise artifacts, 
    while closing bridges fragmented crack pixels and fills small gaps.
    
    \item \textbf{Contour extraction and filtering}: Extract external contours 
    from the binary mask and filter by two criteria: (i) area must be between
    91 pixels and 10\% of the image area, to exclude small areas of noise and very large non-crack regions, and (ii) aspect ratio must be $\geq 1.5$, to prioritize 
    elongated crack-like features over round or square artifacts.
\end{enumerate}

Both methods were evaluated on a common set of pixel-level and 
object-level metrics: recall, precision, F1 score, specificity, intersection over 
union (IoU), and total crack count. These quantities are defined in 
Eqs.~\eqref{eqn:recall}--\eqref{eqn:iou}, where TP denotes a true 
positive (a crack pixel correctly classified as crack), FN denotes a false negative (a crack pixel incorrectly classified as non-crack), FP denotes a false positive (a non-crack pixel incorrectly classified as crack), and TN denotes a true negative (a non-crack pixel correctly classified as non-crack).

\begin{equation}\label{eqn:recall}
    \text{Recall} = \frac{TP}{TP+FN}
\end{equation}
\begin{equation}\label{eqn:precision}
    \text{Precision} = \frac{TP}{TP+FP}
\end{equation}
\begin{equation}\label{eqn:F1Score}
    \text{F1 score} = \frac{2 \cdot \text{Precision} \cdot \text{Recall}}{\text{Precision} + \text{Recall}}
\end{equation}
\begin{equation}\label{eqn:Specificity}
    \text{Specificity} = \frac{TN}{TN+FP}
\end{equation}
\begin{equation}\label{eqn:iou}
    \text{IoU} = \frac{TP}{TP+FP+FN}
\end{equation}

The morphological baseline pipeline was 
tuned to provide the best crack detection performance on the training and validation images, measured with F1 score and IoU, and then tested on the test image subset. 780 distinct parameter combinations were evaluated systematically, including variations
in the Otsu threshold scaling factor, closing iterations, and the
minimum contour area. Across these combinations, F1 score ranged from 0.26 to 0.65 on the training set, with recall exhibiting particularly high sensitivity (0.17--0.58), reflecting the strong dependence of threshold-based detection on the choice of thresholding cutoff. The Otsu scaling factor was the dominant driver of this variability. Varying it across its tested range produced a mean F1 spread of 0.33 and a mean recall spread of 0.35, while the number of closing iterations produced a secondary spread of 0.04 in F1 and the minimum contour area had a negligible effect (spread $<$ 0.01). This range underscores that the performance of classical morphological pipelines is not stable by default and requires deliberate tuning to achieve competitive results. This morphological approach represents a
realistic baseline, because a domain expert would reasonably invest some effort
in tuning classical image processing parameters to improve results before
considering more complex approaches. The resulting pipeline thus reflects 
the state-of-the-art of traditional morphological methods when applied 
by a motivated practitioner.

YOLOEZ model training convergence is characterized by segmentation 
loss and mask-level validation metrics, the training curves for which are 
presented in Section~\ref{sec:results}. Segmentation loss is computed as 
the binary cross-entropy between the predicted mask $\hat{M}$ and the 
ground-truth mask $M$ over all pixels $i$, where $M_i, \hat{M}_i \in \{0,1\}$:
\begin{equation}\label{eqn:seg_loss}
    \mathcal{L}_{\text{seg}} = -\frac{1}{N}\sum_{i=1}^{N}
    \left[ M_i \log \hat{M}_i + (1 - M_i)\log(1 - \hat{M}_i) \right]
\end{equation}

Mask-level detection quality is summarized by the mean average precision 
at IoU threshold 0.5, denoted mAP50(M), which is computed as the 
area under the precision-recall curve evaluated on predicted segmentation 
masks at an IoU threshold of 0.5:
\begin{equation}\label{eqn:map50}
    \text{mAP50(M)} = \int_{0}^{1} P(R)\, dR \Big|_{\text{IoU} \geq 0.5}
\end{equation}
where $P(R)$ is the precision-recall curve evaluated on predicted 
segmentation masks for the crack class.

\section{Results}
\label{sec:results}

\subsection{Qualitative Comparison}
Table~\ref{tab:YOLO_comparison} presents the results of a qualitative analysis comparing YOLOEZ, Ultralytics, Roboflow, and PyTorch as tools for labeling data, training models, and performing inference with a YOLO-based computer vision system. A checkmark ($\checkmark$) indicates that the feature is present, while an $\times$ indicates it is absent. The GUI category indicates whether the platform provides an integrated graphical user interface for dataset management, training configuration, and visualization of results. Data Labeling refers to built-in annotation tools for creating and managing bounding box and segmentation labels. Model Training denotes native support for training YOLO models within the framework. Local Data indicates whether datasets and models can be managed and processed entirely on a local machine without requiring cloud-based storage or services. Inference reflects the ability to deploy trained models for prediction on new images. Open Source specifies whether the core framework is publicly available and modifiable under an open-source license. Code Free identifies tools that do not require programming knowledge to perform training or inference tasks. Multi-model Support indicates whether the platform can handle multiple model architectures or variants, or manage multiple models in parallel. Cloud-based Training denotes frameworks that allow training on remote servers rather than exclusively locally. Fully Free indicates whether all features of the platform are available at no cost, with no subscription or payment required for any functionality. Finally, Multi-class Support indicates whether the platform can train and deploy models that detect multiple object classes simultaneously.

\begin{table}[h]
    \centering
    \setlength{\tabcolsep}{3pt}
    \caption{Qualitative comparison of YOLO training tools}
    \renewcommand{\arraystretch}{1.5}
    \small  
    \begin{tabular}{c c c c c}
        \toprule
        Category & YOLOEZ & Ultralytics & Roboflow & PyTorch \\
        \midrule
        GUI & $\checkmark$ & $\checkmark$ & $\checkmark$ & $\times$ \\
        Data Labeling & $\checkmark$ & $\times$ & $\checkmark$ & $\times$ \\
        Model Training & $\checkmark$ & $\checkmark$ & $\checkmark$\tablefootnote{\label{fn:paid}Only available in the paid tier.} & $\checkmark$ \\
        Local Data & $\checkmark$ & $\checkmark$ & $\times$ & $\checkmark$ \\
        Inference & $\checkmark$ & $\checkmark$ & $\checkmark$ & $\checkmark$ \\
        Open Source & $\checkmark$ & $\checkmark$\tablefootnote{\label{fn:prop}Some proprietary features are not open-sourced.} & $\checkmark$\textsuperscript{\ref{fn:prop}} & $\checkmark$ \\
        Code Free & $\checkmark$ & $\times$ & $\checkmark$ & $\times$ \\
        \makecell{Multi-model \\ Support} & $\checkmark$ & $\checkmark$ & $\checkmark$ & $\checkmark$ \\
        \makecell{Cloud-based \\ Training} & $\times$ & $\checkmark$ & $\checkmark$\textsuperscript{\ref{fn:paid}} & $\times$ \\
        Fully Free & $\checkmark$ & $\times$ & $\times$ & $\checkmark$ \\
        \makecell{Multi-class \\ Support} & $\times$ & $\checkmark$ & $\checkmark$ & $\checkmark$ \\
        \bottomrule
    \end{tabular}
    \label{tab:YOLO_comparison}
\end{table}

\subsection{Quantitative Comparison}
Table~\ref{tab:quant_results} presents the pixel-level and object-level 
performance metrics for YOLOEZ and the morphological baseline across the 
15 test images. Some images in the test set did not include cracks. These images were not used in the calculation of average recall, precision, F1 score, or IoU, because these metrics are undefined or uninformative when no ground-truth cracks are present.

\begin{table}[h]
    \centering
    \setlength{\tabcolsep}{6pt}
    \caption{Quantitative comparison of YOLOEZ and morphological crack detection}
    \small  
    \begin{tabular}{c c c c}
        \toprule
        Metric & \makecell{YOLOEZ \\ ($\pm 95\%$ CI)} & Morphological & \makecell {Ground\\Truth} \\
        \midrule
        Recall         & \textbf{0.6065 $\pm$ 0.032} & 0.4372 & N/A \\
        Precision      & 0.5820 $\pm$ 0.044 & \textbf{0.7403} & N/A \\
        F1 score       & \textbf{0.5591 $\pm$ 0.027} & 0.5353 & N/A \\
        IoU            & \textbf{0.4150 $\pm$ 0.024} & 0.3962 & N/A \\
        Specificity    & 0.9934 $\pm$ 0.002 & 0.9951 & N/A \\
        \makecell{Total cracks \\ detected} & \makecell{\textbf{84 $\pm$ 10}} & \makecell{129} & \makecell{63} \\
        \bottomrule
    \end{tabular}
    \label{tab:quant_results}
\end{table}

\begin{figure}[h]
\centering
\includegraphics[width=0.45\textwidth]{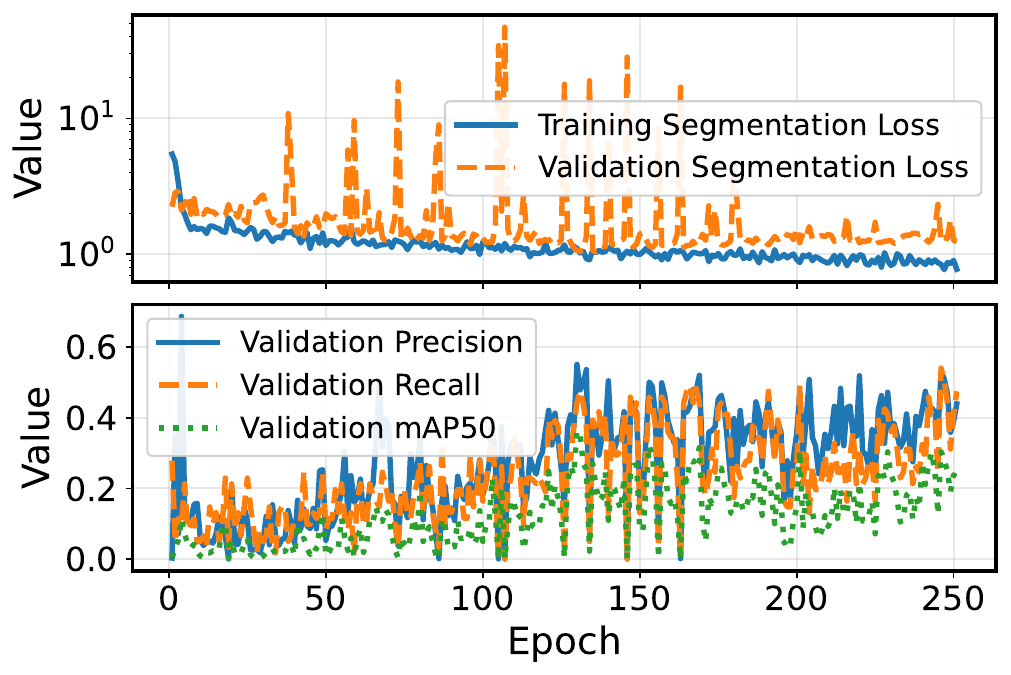}
\caption{Training and validation loss curves for YOLOEZ over 250 epochs (top), and evolution of validation precision, recall, and mAP50(M) (bottom). The relatively small gap between training and validation loss indicates minimal overfitting despite the small training set size of 20 labeled images.}
\label{fig:training_curves}
\end{figure}

Figure~\ref{fig:training_curves} shows a representative sample of the training and validation loss 
curves for YOLOEZ over 250 epochs, along with the evolution of 
precision, recall, and mAP50(M) on the validation set, from one of the ten model training runs. The best checkpoint, selected by the highest recorded mAP50(M) on the validation set, was used for all reported test evaluations.

\section{Discussion}
\subsection{Qualitative Comparison}
The qualitative comparison in Table~\ref{tab:YOLO_comparison} highlights 
a fundamental tradeoff between accessibility and flexibility across the 
evaluated tools. YOLOEZ occupies a distinct niche: it is the only 
platform that combines a graphical user interface, built-in data 
labeling, local data management, and code-free operation in a single 
tool. This combination is particularly relevant for SHM researchers who 
possess domain expertise in materials characterization but limited 
experience with machine learning workflows. By removing the need for 
programming knowledge, YOLOEZ lowers the barrier to entry for deploying 
deep learning-based inspection pipelines in laboratory settings where 
software development resources are limited.

Ultralytics and PyTorch, by contrast, offer considerably greater 
flexibility in model architecture selection, training configuration, and 
integration with broader research codebases. However, both require 
programming knowledge that presents a meaningful barrier to adoption 
among domain experts who are not also software practitioners. Ultralytics partially addresses this through its graphical interface, known as Ultralytics HUB, but 
advanced features and cloud training quotas are locked behind a paid 
subscription, which may be prohibitive for academic research groups. 
PyTorch remains the most flexible option and is well-suited for 
researchers who wish to implement custom training loops or integrate YOLO 
models into larger simulation or analysis pipelines, but it offers no 
native data labeling or model management tooling and requires the user to write code for and assemble these components independently.

Roboflow provides a compelling annotation and dataset management 
experience and is widely used in the computer vision community for 
preparing training data. Its labeling interface is mature and supports a 
variety of annotation formats. However, its dependence on cloud 
infrastructure means that raw image data must be uploaded to external 
servers, which raises data governance concerns for research involving 
proprietary or sensitive materials characterization data. Furthermore, 
model training in Roboflow is available only through its paid tier, 
limiting its utility as a fully free solution for academic users.

The absence of multi-class support in YOLOEZ is worth acknowledging as 
a current limitation. For the single-class crack detection task considered 
here, this constraint has no practical impact. However, future SHM 
applications may require simultaneous detection of multiple defect types, such as cracks, voids, and delaminations, in which case a more 
flexible platform would be necessary. This represents a natural direction 
for future development of the YOLOEZ tool. Taken together, the 
qualitative comparison suggests that YOLOEZ addresses a genuine gap in 
the existing ecosystem for domain-expert-driven SHM applications, 
prioritizing ease of use and data locality at the expense of some 
configurability.

The accessibility of YOLOEZ has direct implications for digital twin workflows, where continuous or periodic visual inspection data must be processed and integrated into a live structural model. By enabling domain experts to develop and deploy inspection models without programming expertise, YOLOEZ lowers the barrier to generating the labeled defect data that feeds these frameworks. Similarly, in predictive maintenance contexts where crack propagation must be tracked over time, the ability to rapidly retrain and redeploy models as new inspection data becomes available is a practical advantage of YOLOEZ's integrated workflow.

\subsection{Quantitative Comparison}
The quantitative results in Table~\ref{tab:quant_results} demonstrate
that YOLOEZ outperformed the morphological baseline across all metrics
except precision and specificity, despite being trained on only 20 labeled images. The
95\% confidence intervals, derived from ten independent training runs,
indicate that this performance advantage is consistent and not an artifact
of a single favorable initialization.

\subsubsection{Recall and Crack Detection Coverage}
YOLOEZ achieved substantially higher recall than the morphological
baseline (0.6065 $\pm$ 0.032 vs. 0.4372), indicating that it
successfully identified a greater fraction of true crack pixels across
the test set. In the SHM context, recall is arguably the most
safety-critical metric: a missed crack detection could result in an
undetected structural defect progressing to failure, whereas a false
positive merely triggers unnecessary follow-up inspection. The
morphological method's lower recall reflects a fundamental limitation of
threshold-based approaches, which is the fact that in images where crack contrast is low or
crack morphology deviates from the assumptions embedded in the pipeline
parameters, the method fails to produce detections entirely. YOLOEZ,
having learned crack appearance directly from annotated examples,
demonstrates greater robustness to these variations, and the narrow
confidence interval on its recall suggests this robustness is reliable
across training runs.

\begin{figure}[htbp]
    \centering
    \includegraphics[width=0.4\textwidth]{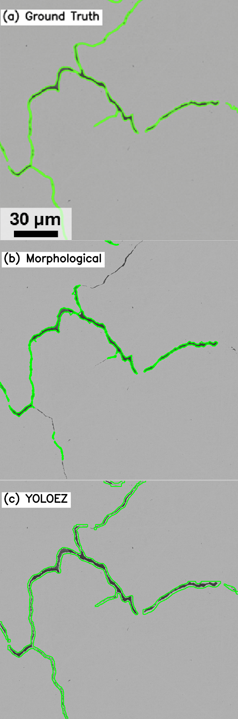}
    \caption{Representative detection outputs on an in-distribution SEM test image, with detected crack boundaries shown in green. (a) Ground truth. (b) Morphological baseline. (c) YOLOEZ.}
    \label{fig:qual_comparison_1}
\end{figure}

\begin{figure}[htbp]
    \centering
    \includegraphics[width=0.4\textwidth]{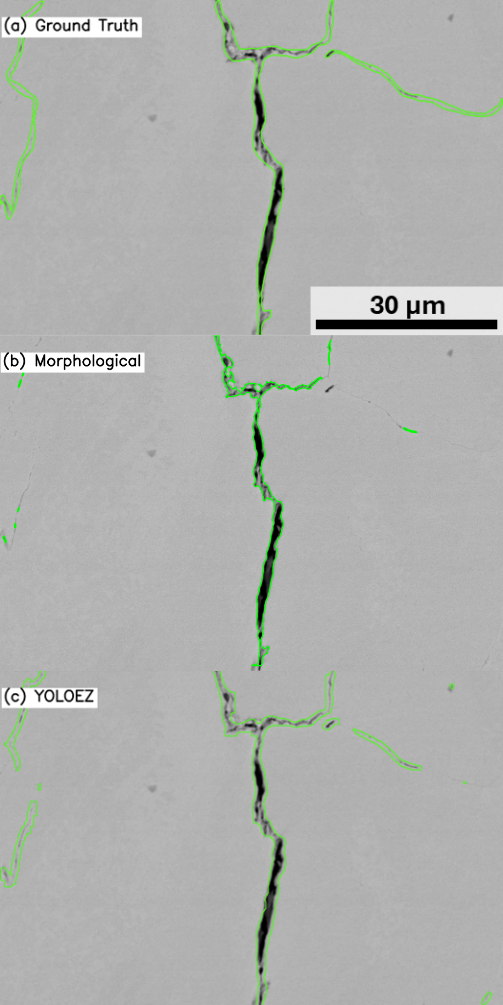}
    \caption{Representative detection outputs on an out-of-distribution SEM test image from a separate tungsten sample, with detected crack boundaries shown in green. (a) Ground truth. (b) Morphological baseline. (c) YOLOEZ.}
    \label{fig:qual_comparison_2}
\end{figure}

Representative detection outputs for both methods on selected test images
are shown in Fig.~\ref{fig:qual_comparison_1} and
Fig.~\ref{fig:qual_comparison_2}, with crack contours highlighted in green, illustrating the qualitative
differences in detection behavior discussed here.

\subsubsection{Precision, Specificity, and False Positive Behavior}
The morphological baseline achieved higher precision than YOLOEZ (0.7403
vs. 0.5820 $\pm$ 0.044), reflecting its conservative detection behavior.
When the morphological pipeline does flag a region, it tends to
correspond to a genuine crack, because thresholding and contour filtering
suppress low-contrast features. However, this conservatism comes at the
direct cost of recall, as discussed above.

Despite its lower precision, YOLOEZ achieved high specificity
(0.9934 $\pm$ 0.002), and given that the morphological baseline value of
0.9951 falls within this confidence interval, there is no statistically
meaningful difference in specificity between the two methods.
Specificity measures the fraction of true non-crack pixels correctly
identified as background, and the effectively identical values confirm
that neither approach produces widespread false positives across
background regions. 

The object-level crack count provides additional context. The
morphological method detected 129 objects against a ground-truth count
of 63, indicating that while individual detected pixels are often
correct, the method over-segments the image by fragmenting
single cracks into several contours and flagging surface texture artifacts
as spurious detections. YOLOEZ detected 84 $\pm$ 10 objects, more
closely matching the ground-truth count, suggesting substantially better
instance-level detection. An inspector reviewing morphological results
would be confronted with roughly twice the ground-truth count, with a substantial fraction being spurious, making manual review burdensome. YOLOEZ's output presents a
far more manageable set of candidates that more faithfully reflects the
true defect population.

\subsubsection{Localization Quality}
YOLOEZ achieved an IoU of 0.4150 $\pm$ 0.024 compared to 0.3962 for
the morphological method, indicating better spatial
correspondence with ground-truth crack regions. IoU is a particularly
demanding metric for thin, elongated features such as micro-scale cracks,
because small lateral offsets between predicted and ground-truth contours
can reduce the score significantly even when the overall crack trajectory
is correctly identified. The fact that YOLOEZ consistently achieves higher
IoU than the morphological baseline suggests that the
learned segmentation masks capture crack geometry more faithfully than
threshold-derived contours.

\subsubsection{Out-of-Distribution Generalization}
Ten of the 15 test images were drawn from a separate 3D-printed tungsten
sample not seen during training, providing a direct assessment of
out-of-distribution generalization. This is a particularly challenging
scenario, as the new sample may exhibit different surface texture
characteristics, crack morphologies, and imaging conditions relative to
the training data. The representative detection example in
Fig.~\ref{fig:qual_comparison_2} specifically illustrates performance on
an out-of-distribution image, and demonstrates the ability of YOLOEZ to maintain
meaningful detection performance on the out-of-distribution data with confidence
intervals that remain tight. This performance suggests the model has learned features that
generalize beyond the specific sample used for training.

\subsubsection{Training Dynamics and Implications}
The training curves in Fig.~\ref{fig:training_curves} show that
segmentation loss decreased overall over 250 epochs on both the training
and validation sets, with precision, recall, and mAP50(M) generally increasing
throughout. The relatively small gap between training and validation loss
indicates that significant overfitting did not occur, which is notable
given the small training set size. The variance visible in the validation
curves is a consequence of the small validation set of five images, where
a single image can meaningfully shift aggregate metrics between epochs.

\section{CONCLUSION}
This work presented YOLOEZ, an open-source, GUI-based platform
for end-to-end YOLO model development designed to lower the barrier
to AI-driven structural defect detection for domain experts without
programming expertise. A qualitative comparison against existing tools
demonstrated that YOLOEZ is the only platform combining a graphical
interface, built-in data labeling, local data management, and code-free
operation in a single workflow, addressing a genuine gap in the current
ecosystem for SHM applications. Quantitative evaluation on micro-scale
crack detection in SEM images of additively manufactured tungsten
confirmed that YOLOEZ outperforms a classical morphological baseline
across recall, F1 score, IoU, and crack count accuracy, despite being
trained on only 20 labeled images. These results, validated across ten
independent training runs, demonstrate that accessible, GUI-driven deep
learning tools can achieve meaningful detection performance under the
data-constrained conditions typical of SHM research. Future work will
focus on extending YOLOEZ to support multi-class defect detection and
evaluating its performance across a broader range of structural materials
and imaging modalities.

\section{Code Availability}
YOLOEZ is available as an open-source project at: \\
\url{https://github.com/michaelholm6/YOLOEZ}

A detailed software description and documentation are currently being prepared for submission to the Journal of Open Source Software (JOSS).

\section*{Acknowledgment}
GL would like to thank the support of National Science Foundation (DMS-2533878, DMS-2053746, DMS-2134209, ECCS-2328241, CBET-2347401 and OAC-2311848), and U.S.~Department of Energy (DOE) Office of Science Advanced Scientific Computing Research program DE-SC0023161, the SciDAC LEADS Institute, and DOE–Fusion Energy Science, under grant number: DE-SC0024583.
\bibliographystyle{asmeconf}  
\bibliography{asmeconf-sample}

\end{document}